\documentclass[10pt]{article}

\usepackage[letterpaper,margin=1in,headheight=14pt]{geometry}
\usepackage{iftex}
\usepackage{amsmath,amsthm}
\ifPDFTeX
  \input{glyphtounicode}\pdfgentounicode=1
  \usepackage[T1]{fontenc}
  \usepackage[utf8]{inputenc}
  \usepackage{amssymb}
  \IfFileExists{newtxtext.sty}{\usepackage{newtxtext,newtxmath}}{\usepackage{mathptmx}}
  \usepackage[scaled=0.92]{helvet}
  
\else
  \usepackage{fontspec}
  \usepackage{unicode-math}
\fi
\usepackage{graphicx}
\usepackage{booktabs}
\usepackage{enumitem}
\usepackage{titlesec}
\usepackage{fancyhdr}
\usepackage[round,authoryear]{natbib}
\usepackage[hidelinks]{hyperref}

\newif\ifreview
\reviewfalse

\ifreview
  
\else
  
\fi
\titleformat{\section}{\sffamily\bfseries\large}{\thesection}{1em}{}
\titleformat{\subsection}{\sffamily\bfseries\normalsize}{\thesubsection}{1em}{}
\titlespacing*{\section}{0pt}{2.2ex plus .5ex minus .2ex}{1.0ex plus .2ex}
\titlespacing*{\subsection}{0pt}{1.8ex plus .4ex minus .2ex}{0.6ex plus .2ex}
\titleformat{\paragraph}[runin]{\normalfont\bfseries}{}{0pt}{}
\titlespacing*{\paragraph}{0pt}{1.0ex plus .3ex minus .2ex}{0.6em}

\theoremstyle{plain}
\newtheorem{theorem}{Theorem}
\newtheorem{proposition}{Proposition}
\newtheorem{lemma}{Lemma}
\newtheorem{corollary}{Corollary}
\theoremstyle{remark}
\newtheorem*{remark}{Remark}

\newcommand{\qh}{\hat q}
\newcommand{\kh}{\hat k}
\newcommand{\WQ}{W_Q}
\newcommand{\WK}{W_K}
\newcommand{\dn}{\tilde\nabla}
\newcommand{\ip}[2]{\langle #1,\, #2\rangle}
\newcommand{\norm}[1]{\lVert #1\rVert}
\newcommand{\R}{\mathbb{R}}
\newcommand{\QR}{\mathrm{Q_R}}
\newcommand{\QTU}{\mathrm{Q_{TU}}}

\newcommand{\EmailZhang}{yuanqizhang375@gmail.com}
\newcommand{\EmailWang}{hcwang\_28@seu.edu.cn}
\newcommand{\AffilWang}{School of Mathematics, Southeast University, Nanjing, China}
\newcommand{\EmailLin}{congpinglin@hust.edu.cn}
\newcommand{\authorentry}[3]{%
  {\raggedright\bfseries #1\hfill\normalfont\small\texttt{#2}\par}%
  {\raggedright\small #3\par}\vspace{6pt}}

\begin{document}

\thispagestyle{fancy}
\vspace*{0.2in}
{\raggedright\sffamily\bfseries\LARGE
The Gradient Difference Induced by Detaching the Denominator in QK Normalization\par}
\vspace{0.3in}
\ifreview
  {\raggedright\bfseries Anonymous authors\par}
  {\raggedright Paper under double-blind review\par}
\else
  \authorentry{Yuanqi Zhang\textsuperscript{*}}{\EmailZhang}{School of Mathematics and Statistics, Huazhong University of Science and Technology, Wuhan, China}
  \authorentry{Haocheng Wang\textsuperscript{*}}{\EmailWang}{\AffilWang}
  \authorentry{Congping Lin\textsuperscript{\dag}}{\EmailLin}{School of Mathematics and Statistics, Huazhong University of Science and Technology, Wuhan, China}
  {\raggedright\small \textsuperscript{*}Equal contribution. \quad \textsuperscript{\dag}Corresponding author.\par}
\fi
\vspace{0.3in}

\begin{center}
{\sffamily\bfseries\large Abstract}
\end{center}
\begin{list}{}{\setlength{\leftmargin}{0.5in}\setlength{\rightmargin}{0.5in}\setlength{\parsep}{5.5pt}}
\item\relax
QK normalization divides each query and key by its own RMS norm before the
attention scores are computed. Treating this denominator as a constant during
backpropagation leaves the forward pass unchanged but alters the parameter
gradients. We derive a closed form for the difference between the gradient
under this detached rule and the gradient under the standard rule. For a
single token, this difference is parallel to the token's own query; it changes
only the query norm and has no first-order effect after normalization. The
queries, however, are computed by a projection matrix $\WQ$ shared across all
tokens. The gradient difference on $\WQ$ therefore equals $\WQ$
right-multiplied by a weighted second moment of the inputs, and it decomposes
exactly into three components: a radial component that changes only the norm
of $\WQ$, a tangential component that changes the query directions of other
tokens, and an upstream component that changes the gradient passed to earlier
layers. Under AdamW, the radial component drives $\WQ$ to a larger equilibrium
norm and reduces its angular update per step. Paired experiments on a small
Transformer language model are consistent with this analysis. Detaching the
query denominator consistently raises the test loss, and the gap persists
across hardware, evaluation datasets and learning rates. All three components
remain non-negligible throughout training, with the tangential component the
largest. Norm growth and the smaller angular update appear only on the
detached projection matrix, while the other projection matrix is essentially
unchanged. With a learnable scale, however, we observe no resolvable loss gap.
\end{list}

\section{Introduction}

Large language models are now used in conversational assistants, programming
tools, machine translation and scientific research. Training them consumes
substantial compute, and both the cost and the final quality depend on whether
training is stable and whether each update is effective. It is therefore worth
understanding what a seemingly harmless change to the training procedure
actually does.

Nearly all current large language models are built on the Transformer
\citep{vaswani2017}. Its attention mechanism lets every token gather
information from other positions: the vector of each token is multiplied by
matrices shared across all tokens to produce a query and a key; the inner
product of one token's query with another token's key gives an attention
score, and the scores in a row are turned into weights by a softmax.

The norms of the queries and keys directly set the magnitude of the scores,
and training by itself does not constrain these norms. QK normalization
\citep{henry2020} divides each query and each key by its own norm before the
attention scores are computed. The magnitude of the scores then no longer
grows with the query and key norms, and the instabilities caused by
uncontrolled growth of attention scores in large-scale training are mitigated
\citep{dehghani2023,wortsman2024}.

This paper studies an intervention on QK normalization: the forward pass
divides by the norm as usual, while backpropagation treats this norm as a
constant that does not depend on the query. Every quantity computed in the
forward pass is unchanged; only the gradients differ. Treating normalization
statistics in this way is not unusual. It appears in the analysis of LayerNorm
\citep{xu2019}, in interpretability methods \citep{ali2022,ameisen2025} and
in training recipes \citep{loshchilov2026}.

A common intuition is that this intervention should be harmless: normalization
depends only on direction, the derivative that passes through the denominator
concerns only the norm, and the norm has already been divided out in the
forward pass. We ran a paired experiment on a 6-layer Transformer with
17.65 million parameters. Detaching the query denominator in all layers raises
the test loss by 0.060 nats per token relative to standard training. Why
training degrades when the forward pass is identical and only the backward
pass changes is the question this paper answers.

Prior work has described the geometry of scale-invariant weights under
gradient descent and has reported that detaching normalization statistics in
the backward pass harms training. It has not analyzed the role that the
derivative through the normalization denominator plays in the parameter
update, nor given a closed form for how this term acts on other tokens through
the shared projection matrix.

\paragraph{QK normalization.}
Existing work is concerned with the numerical stability and expressivity of
QK normalization in the forward pass
\citep{henry2020,dehghani2023,wortsman2024,tao2026}. \citet{zhuo2025} analyze
the gradients of the weights under QK and QKV normalization using the full
Jacobian of the normalization, and bound their norms. We ask instead how a
change to the backward pass alters training when the forward pass is fixed.

\paragraph{Freezing normalization statistics in the backward pass.}
When introducing batch normalization, \citet{ioffe2015} already noted that if
the gradient step ignores the dependence of the normalization on the
parameters, the updated parameter grows indefinitely while the loss stays
fixed. DetachNorm of \citet{xu2019} treats the mean and variance of LayerNorm
as constants in the backward pass; they find that freezing the variance
degrades machine-translation training, to the point of divergence, and
attribute this, on a single vector, to the derivative of the variance
rescaling the gradients. Interpretability methods linearize normalization in
the backward pass \citep{ali2022,ameisen2025}, and some training recipes use
modified or approximate gradients in the backward pass
\citep{shen2020,loshchilov2026}. These works examine the forward output or
the final loss. Our object is QK normalization inside attention, and we write
the derivative through the denominator on the shared projection matrix, which
gives the closed form $\WQ M$.

\paragraph{Scale invariance and the effective learning rate.}
Normalization makes the output invariant to the scale of the weights, so the
gradient is orthogonal to the weights and the effective learning rate
decreases as the weights grow
\citep{vanlaarhoven2017,hoffer2018,arora2019,wan2021,kunin2021}. The absence
of a radial component in the standard gradient, stated in our
Theorem~\ref{thm:decomp}(i), is a special case of these results, and the
detached rule breaks it. \citet{kosson2024} analyze the rotational
equilibrium between weight norm and angular update under AdamW; the analysis
in Section~\ref{sec:adamw} follows their approach and adds the nonzero cosine
caused by the radial component. \citet{prieto2025} observe that after
overfitting the gradient aligns with the weight-rescaling direction that only
enlarges the logits, and propose removing the gradient component parallel to
the weights; the direction that scales a row of attention scores uniformly in
our Lemma~\ref{lem:beta} is analogous.

Building on these works, we write in closed form how this derivative acts in
QK normalization, and use it to answer the question above: the intuition
holds for a single token and fails for shared parameters. For a single token,
the gradient difference is parallel to its own query; if the query were
itself a parameter, the normalized queries after one step under the two rules
would differ only at second order in the learning rate
(Corollary~\ref{cor:free}). The queries, however, are computed by a matrix
$\WQ$ shared across all tokens, and the per-token gradient differences add up
on $\WQ$ to $\WQ$ right-multiplied by a weighted second moment $M$ of the
inputs (Proposition~\ref{prop:closed}); the term that only changes the norm
for token $i$ rotates the query direction of another token $j$ at first order
(Corollary~\ref{cor:rot}). We further decompose the gradient difference
exactly into radial, tangential and upstream components, give necessary and
sufficient conditions for the tangential and upstream components to vanish,
and express the detached rule as the standard gradient of an auxiliary
objective.

The paper is organized as follows. Section~\ref{sec:setup} sets up the problem
and notation. Sections~\ref{sec:closed} and~\ref{sec:decomp} derive the closed
form of the gradient difference and its decomposition into three components.
Section~\ref{sec:adamw} analyzes the consequences of the radial component
under AdamW. Section~\ref{sec:exp} tests these conclusions in training
experiments, and Section~\ref{sec:conclusion} concludes.

\section{Problem Setup and Notation}
\label{sec:setup}

We consider a single attention head in one layer. The outputs of the preceding
layers pass through an RMSNorm \citep{zhang2019} to give the input of this
head. We denote the input of token $i$ by $x_i \in \R^m$, where $m$ is the
model width. The projection matrices $\WQ, \WK \in \R^{d\times m}$ are shared
by all tokens of the head, where $d$ is the head dimension. Queries and keys
are defined as
\begin{equation*}
q_i = \WQ\, x_i, \qquad k_j = \WK\, x_j .
\end{equation*}
QK normalization divides each query by its RMS norm $s_i$,
\begin{equation*}
s_i = \sqrt{\norm{q_i}^2 / d + \varepsilon}, \qquad \qh_i = \frac{q_i}{s_i} ,
\end{equation*}
where $\varepsilon$ is a small constant that prevents division by zero, set to
$10^{-5}$ in our experiments. Keys are normalized in the same way to $\kh_j$,
and the derivation for keys is identical, so we state everything for queries.
To keep the derivation clean, we set $\varepsilon = 0$ from here on. Then
$\qh_i$ is invariant to positive rescaling of $q_i$, so multiplying $\WQ$ by a
positive scalar changes none of the $\qh_i$ of this head and leaves the
network output unchanged; we refer to this as scale invariance. The case
$\varepsilon > 0$ is discussed in Appendix~\ref{app:eps}.

The attention score of token $i$ on token $j$ is
\begin{equation*}
z_{ij} = \frac{\ip{\qh_i}{\kh_j}}{\sqrt d} .
\end{equation*}
The scores in a row are turned into attention weights by a softmax over the
positions visible to token $i$ (causal mask). The loss $L$ is the mean
negative log-probability of the correct token over all positions in a batch;
the architecture is described in Section~\ref{sec:exp-setup}.

We next introduce notation for the gradients that enter the chain rule. We
write $\delta_i = \partial L/\partial \qh_i$ for the backpropagated gradient
\citep[notation following][]{rumelhart1986}. It is the gradient that reaches
$\qh_i$ from everything computed after $\qh_i$, namely the attention scores,
the rest of this layer, and all later layers up to the output. We write
$J_i = \partial \qh_i/\partial q_i$, which is (see Appendix~\ref{app:jac} for
the derivation)
\begin{equation}
J_i = \frac{1}{s_i}\Bigl(I - \frac{q_i q_i^\top}{\norm{q_i}^2}\Bigr).
\label{eq:jac}
\end{equation}
The gradient of the loss with respect to $q_i$ is therefore
$J_i^\top\delta_i$.

We now introduce the detached rule \citep{xu2019,ali2022} and analyze how it
changes the gradients in QK normalization. This device is already widely used
with other normalizations. It keeps the forward pass as it is and treats $s_i$
as a constant when differentiating (a stop-gradient). This changes the
gradient quantities introduced above, and this change is the object of our
study. For any variable $v$, we write $\nabla_v L$ for the gradient of $L$
with respect to $v$ under the standard rule and $\dn_v L$ for the gradient
under the detached rule, and we study the gradient difference
$\dn_v L - \nabla_v L$. Below, $v$ is taken to be $q_i$, $\WQ$ or $x_i$,
corresponding to different stages of the analysis.

Under the detached rule, the derivative of $\qh_i$ with respect to $q_i$
becomes $I/s_i$ in place of $J_i$, and the gradient of the loss with respect
to $q_i$ becomes $\delta_i/s_i$. The difference between the two rules, that
is, the term by which $J_i$ and $I/s_i$ differ, is exactly the derivative that
passes through $s_i$. The effect on the loss of everything after $\qh_i$ is
contained in $\delta_i$, so the results below do not depend on the downstream
architecture or on the loss.

We briefly outline the following sections. The step $\qh_i \to q_i$ is where
the two rules differ; the step $q_i \to \WQ$ sums over all tokens that share
$\WQ$ (Section~\ref{sec:closed}); and the step $q_i \to x_i$ passes the
gradient difference to earlier layers (Section~\ref{sec:decomp}). Throughout,
we compare the gradients of the two rules, and the resulting one-step updates,
at the same parameters and on the same batch. The loss gap accumulated over
many steps is measured in the experiments of Section~\ref{sec:exp}.

\section{Closed Form of the Gradient Difference}
\label{sec:closed}

This section and the following ones study the gradient difference, that is,
the gradient under the detached rule minus the gradient under the standard
rule. The standard rule serves as the reference.

\subsection{Per-Token and Shared-Parameter Forms}
\label{sec:forms}

\begin{lemma}\label{lem:beta}
Fix a token $i$ and multiply the scores $z_{ij}$ of row $i$ by a positive
scalar $\beta$ for all $j$. The derivative of the loss at $\beta = 1$ is
\begin{equation*}
\frac{\mathrm{d} L}{\mathrm{d} \beta}\Big|_{\beta=1}
 = \sum_j z_{ij}\, \frac{\partial L}{\partial z_{ij}}
 = \ip{\qh_i}{\delta_i} .
\end{equation*}
\end{lemma}

\begin{proof}
The first equality is the chain rule. By the definition of the scores, $\qh_i$
appears only in row $i$, and $\delta_i$ is the sum over $j$ of
$\partial L/\partial z_{ij}$ times $\kh_j/\sqrt d$. Taking the inner product
with $\qh_i$ gives the second equality.
\end{proof}

\begin{remark}
The value $\beta = 1$ corresponds to the original, unscaled network. The lemma
describes how the loss changes when $\beta$ is moved slightly away from $1$,
starting from the original network: it gives the first-order rate of change of
the loss when row $i$ of the scores is scaled up slightly. Heuristically,
multiplying a row of scores by $\beta$ amounts to setting the softmax
temperature of that row to $1/\beta$ \citep{hinton2015,guo2017}. A larger
$\beta$ makes the attention distribution of the row more concentrated and
lowers its entropy; the distribution tends to uniform as $\beta \to 0$ and
concentrates on the position with the largest score as $\beta \to \infty$.
Hence $\ip{\qh_i}{\delta_i} < 0$ means that scaling up row $i$ lowers the
loss, that is, the loss favors more concentrated attention in this row.
\end{remark}

\begin{proposition}\label{prop:closed}
Assume $q_i \neq 0$ for all $i$.
\begin{enumerate}[label=(\alph*),leftmargin=2em,topsep=2pt,itemsep=2pt]
\item For each token $i$, the gradient difference on a single token is
\begin{equation*}
\dn_{q_i} L - \nabla_{q_i} L
 = \frac{\ip{\qh_i}{\delta_i}}{\norm{q_i}^2}\; q_i .
\end{equation*}
\item Summed over tokens, the gradient difference on $\WQ$ is
\begin{equation*}
\dn_{\WQ} L - \nabla_{\WQ} L
 = \sum_i \bigl(\dn_{q_i} L - \nabla_{q_i} L\bigr)\, x_i^\top = \WQ M ,
\end{equation*}
where
\begin{equation*}
M = \sum_i \frac{\ip{\qh_i}{\delta_i}}{\norm{q_i}^2}\; x_i x_i^\top
 \in \R^{m\times m} .
\end{equation*}
\item One SGD step with learning rate $\eta$ satisfies
\begin{equation*}
\WQ - \eta\, \dn_{\WQ} L = \WQ\,(I - \eta M) - \eta\, \nabla_{\WQ} L .
\end{equation*}
\end{enumerate}
\end{proposition}

The proof is given in Appendix~\ref{app:prop1}.

\begin{remark}
We state (c) for SGD because its update is the learning rate times the
gradient, with no transformation by the optimizer, so the effect of the
gradient difference on the parameters is seen in isolation.
\end{remark}

We now interpret the three parts of Proposition~\ref{prop:closed}.

Part (a) shows that the gradient difference of each token is parallel to
$q_i$. Rescaling $q_i$ does not change $\qh_i$, and the loss depends on $q_i$
only through $\qh_i$, so $L(c\,q_i) = L(q_i)$ for every $c > 0$.
Differentiating in $c$ at $c = 1$ gives $\ip{q_i}{\nabla_{q_i} L} = 0$: the
standard gradient is orthogonal to $q_i$ and does not change $\norm{q_i}$ at
first order. Since the gradient under the detached rule is the standard
gradient plus the gradient difference, its entire component along $q_i$ comes
from the gradient difference. The sign of the coefficient is that of
$\ip{\qh_i}{\delta_i}$, which by Lemma~\ref{lem:beta} is the rate of change of
the loss when row $i$ of the scores is scaled up. The coefficient is negative
when the angle between $\delta_i$ and $\qh_i$ exceeds $90^\circ$, meaning that
more concentrated attention in this row lowers the loss. In that case the
gradient difference in (a) is a negative multiple of $q_i$. Because the
detached rule treats $s_i$ as a constant, it mistakes lengthening $q_i$ for a
way of scaling up this row of scores, and its gradient-descent step lengthens
$q_i$ relative to the standard rule. This norm is divided out in the forward
pass, however: $\qh_i$ and the scores $z_{ij}$ are unchanged, and so are the
network output and the loss.

Part (b) is the aggregate of (a) on the shared parameter $\WQ$.

Part (c) shows that, compared with the standard rule, one step of the detached
rule carries an extra right factor $I - \eta M$, that is, a multiplicative
rescaling of $\WQ$ along the input directions. $\WQ$ is amplified along
eigendirections of $M$ with negative eigenvalues and shrunk along those with
positive eigenvalues. When the factor is the same in all directions, $\WQ$
only grows as a whole; when it differs across directions, the direction of
$\WQ$ changes as well. Section~\ref{sec:decomp} calls these two parts the
radial and tangential components. In this way, the term that only changes the
norm for token $i$ acts, through the shared $\WQ$, on the queries of other
tokens. Section~\ref{sec:coupling} makes this precise.

\subsection{Cross-Token Coupling Through Shared Parameters}
\label{sec:coupling}

We first consider a hypothetical setting in which the update of one token does
not affect the queries of other tokens: each $q_i$ is an independent parameter
rather than being computed from $\WQ$.

\begin{corollary}\label{cor:free}
Treat each $q_i$ directly as a parameter and take one SGD step with learning
rate $\eta$ under each rule. The resulting $\qh_i$ differ by $O(\eta^2)$.
\end{corollary}

The proofs of Corollaries~\ref{cor:free}--\ref{cor:norm} are given in
Appendix~\ref{app:prop1}.

The intuition that an unchanged forward pass implies no harm holds in this
setting. In an actual network, however, the queries are computed from a shared
$\WQ$, and Proposition~\ref{prop:closed}(c) applies. The gradient difference
acts on the shared $\WQ$ through the right factor $I - \eta M$ and therefore
changes the queries of other tokens. Corollaries~\ref{cor:rot}
and~\ref{cor:norm} give its first-order effect on the direction and on the
norm of the queries.

\begin{corollary}\label{cor:rot}
In the step of Proposition~\ref{prop:closed}(c), the gradient difference
changes the query of token $j$ by
\begin{equation*}
-\eta\, \WQ M\, x_j
 = -\eta \sum_i \frac{\ip{\qh_i}{\delta_i}}{\norm{q_i}^2}\,
   \ip{x_i}{x_j}\; q_i .
\end{equation*}
Since $J_j = \partial \qh_j/\partial q_j$, left-multiplying this change of
$q_j$ by $J_j$ gives the first-order change of $\qh_j$,
\begin{equation*}
-\eta\, J_j \WQ M\, x_j
 = -\frac{\eta}{s_j}\Bigl(I - \frac{q_j q_j^\top}{\norm{q_j}^2}\Bigr)
   \sum_{i \ne j} \frac{\ip{\qh_i}{\delta_i}}{\norm{q_i}^2}\,
   \ip{x_i}{x_j}\; q_i .
\end{equation*}
\end{corollary}

By the second identity of Corollary~\ref{cor:rot}, $\qh_j$ is rotated at first
order as soon as some $q_i$ is not parallel to $q_j$ and
$\ip{x_i}{x_j} \neq 0$. The term $i = j$ is removed by the projection, which
is why this effect is invisible in a single-token analysis. The same term also
changes the norm of each query.

\begin{corollary}\label{cor:norm}
After the same step, the first-order term of the difference in $\norm{q_i}^2$
between the two rules is
\begin{equation*}
-2\eta \sum_j \frac{\ip{\qh_j}{\delta_j}}{\norm{q_j}^2}\,
 \ip{x_i}{x_j}\, \ip{q_i}{q_j} ,
\end{equation*}
in which the term $j = i$ equals
$-2\eta\,\ip{\qh_i}{\delta_i}\,\norm{x_i}^2$.
\end{corollary}

These results describe two effects that arise from the shared $\WQ$ and are
invisible in a single-token analysis. Under the detached rule, the query of
token $j$ receives an additional term, which is a weighted sum of the queries
$q_i$ of the other tokens. The weights are set by the coefficient
$\ip{\qh_i}{\delta_i}/\norm{q_i}^2$ of $q_i$ and by the input inner product
$\ip{x_i}{x_j}$. When the coefficient is negative and the inputs are
positively correlated, $q_j$ is pulled toward $q_i$, so the query directions
of different tokens tend to align. The norm of a query is likewise no longer
determined by the coefficient of its own token alone; it also receives
contributions from other tokens through the shared $\WQ$.

\subsection{An Equivalent Auxiliary Objective}
\label{sec:aux}

\begin{proposition}\label{prop:aux}
Let $j$ range over all detached normalization sites, with $\delta_j$ evaluated
in the backward graph of the detached rule. Then, for all parameters $\theta$,
\begin{equation*}
\dn_\theta L = \nabla_\theta \Bigl[\, L
 + \sum_j \ip{\qh_j}{\delta_j}\, \log s_j \Bigr],
\end{equation*}
where the coefficients $\ip{\qh_j}{\delta_j}$ are treated as constants under
differentiation. The identity also holds for $\varepsilon > 0$.
\end{proposition}

\begin{remark}
We use the index $j$ here rather than $i$ to mark a distinction. Earlier, $i$
ranges only over the tokens of one head. In Proposition~\ref{prop:aux}, $j$
ranges over the detached normalization sites in all layers and all heads,
because the auxiliary objective sums over every detached site in the network.
\end{remark}

The proof, an induction over the detached sites from output to input, is
given in Appendix~\ref{app:aux}.

Proposition~\ref{prop:aux} shows that gradient descent under the detached rule
is equivalent to standard gradient descent on $L$ plus the auxiliary term
$\sum_j \ip{\qh_j}{\delta_j} \log s_j$, with the coefficients
$\ip{\qh_j}{\delta_j}$ held constant. When a coefficient is negative, the
auxiliary term decreases as $s_j$ increases, so a step along its negative
gradient increases $s_j$. But $s_j$ does not enter the forward output:
increasing it changes neither $\qh_j$ nor the attention scores. The gradient
of the auxiliary term with respect to $\WQ$ is the $\WQ M$ of
Proposition~\ref{prop:closed}(b).

\section{Radial, Tangential and Upstream Decomposition}
\label{sec:decomp}

To separate the part of the gradient difference that only changes the norm of
$\WQ$ from the part that changes its direction, we project the $\WQ M$ of
Proposition~\ref{prop:closed}(b) orthogonally onto the direction of $\WQ$ and
onto its orthogonal complement, under the Frobenius inner product
$\ip{A}{B} = \operatorname{tr}(A^\top B)$. This gives the radial component and
the tangential component, which are orthogonal and sum to $\WQ M$:
\begin{equation*}
(\WQ M)_{\parallel} = \frac{\ip{\WQ}{\WQ M}}{\norm{\WQ}^2}\, \WQ, \qquad
(\WQ M)_{\perp} = \WQ M - (\WQ M)_{\parallel} .
\end{equation*}
The gradient difference is also passed to earlier layers through
$q_i = \WQ x_i$, as the gradient difference
$\dn_{x_i} L - \nabla_{x_i} L$ with respect to $x_i$. We call it the upstream
component. Note that it is not part of $\WQ M$: the radial and tangential
components are matrices on $\WQ$, whereas the upstream component is a vector
on each $x_i$. Its closed form is given in Theorem~\ref{thm:decomp}(iii).

\begin{theorem}\label{thm:decomp}
Assume $q_i \neq 0$ for all $i$.
\begin{enumerate}[label=(\roman*),leftmargin=2em,topsep=2pt,itemsep=2pt]
\item The coefficient of the radial component is the sum of
$\ip{\qh_i}{\delta_i}$ over the head, and the standard gradient has no radial
component:
\begin{equation}
\ip{\WQ}{\WQ M} = \sum_i \ip{\qh_i}{\delta_i}, \qquad
\ip{\WQ}{\nabla_{\WQ} L} = 0, \qquad
\ip{\WQ}{\dn_{\WQ} L} = \sum_i \ip{\qh_i}{\delta_i} .
\label{eq:radial}
\end{equation}
\item The first-order change of $\qh_j$ in Corollary~\ref{cor:rot} comes only
from the tangential component. For each token $j$,
\begin{equation*}
J_j\, (\WQ M)_{\parallel}\, x_j = 0, \qquad
J_j\, \WQ M\, x_j = J_j\, (\WQ M)_{\perp}\, x_j .
\end{equation*}
\item When this is the only detached normalization, the gradient difference
with respect to $x_i$ is
\begin{equation}
\dn_{x_i} L - \nabla_{x_i} L
 = \frac{\ip{\qh_i}{\delta_i}}{\norm{q_i}^2}\; \WQ^\top \WQ\, x_i .
\label{eq:upstream}
\end{equation}
\end{enumerate}
\end{theorem}

The proof is given in Appendix~\ref{app:thm1}.

We interpret the three parts in turn. In (i), the standard gradient is
orthogonal to $\WQ$, and the part of the detached-rule gradient along $\WQ$
comes entirely from the radial component, whose coefficient is
$\sum_i \ip{\qh_i}{\delta_i}$ over the head. By Lemma~\ref{lem:beta}, this is
the rate of change of the loss when all attention scores of the head are
scaled up together. Its effect on $\WQ$ is the same as that of the gradient
difference on a single $q_i$ in Section~\ref{sec:forms}: it changes the norm
and not the output, and can therefore affect training only through the
optimizer. Part (ii) states that the first-order effect of the gradient
difference on each $\qh_j$ comes entirely from the tangential component. By
(iii), the upstream component is passed on to the parameters of earlier
layers, including the token embeddings. Since $x_i$ is the output of an
RMSNorm \citep{zhang2019} and the standard rule removes the component along
$x_i$ at that step, whether the upstream component passes through depends on
whether $\WQ^\top \WQ\, x_i$ is parallel to $x_i$.

Proposition~\ref{prop:vanish} gives necessary and sufficient conditions for
the tangential and upstream components to vanish.

\begin{proposition}\label{prop:vanish}
\leavevmode
\begin{enumerate}[label=(\alph*),leftmargin=2em,topsep=2pt,itemsep=2pt]
\item The tangential component vanishes if and only if there is a constant
$\alpha$ such that
\begin{equation*}
M v = \alpha\, v \qquad \text{for every } v \text{ in the row space of } \WQ .
\end{equation*}
\item If $x_i$ is the output of an RMSNorm without gain, the upstream
component vanishes for all tokens after passing through this normalization if
and only if, for every token $i$ with $\ip{\qh_i}{\delta_i} \neq 0$, there is
a constant $\mu_i$ such that
\begin{equation*}
\WQ^\top \WQ\, x_i = \mu_i\, x_i .
\end{equation*}
\end{enumerate}
\end{proposition}

The proof is given in Appendix~\ref{app:vanish}.

In an actual network, neither $M$ nor $\WQ^\top \WQ$ is isotropic. The RMSNorm
in our model has an elementwise gain; the corresponding condition for (b) is
also given in Appendix~\ref{app:vanish}.

The analysis so far concerns the query denominator. As noted in
Section~\ref{sec:setup}, keys are normalized in the same way. If the RMS norm
of the key $k_j$ is instead treated as a constant in the backward pass, which
we call key-side detaching, the same results hold with $\WQ$ replaced by $\WK$
and $\ip{\qh_i}{\delta_i}$ replaced by
$\ip{\kh_j}{\partial L/\partial \kh_j}$ in Proposition~\ref{prop:closed}(b)
and Theorem~\ref{thm:decomp}. The latter equals the sum
$\sum_i z_{ij}\, \partial L/\partial z_{ij}$ over column $j$
(Appendix~\ref{app:key}), whereas Lemma~\ref{lem:beta} is the same sum over
row $i$; the two agree when summed over the head. The query-side coefficients
therefore sit on the positions that attend, and the key-side coefficients on
the positions that are attended to. When all layers are detached
simultaneously, Appendix~\ref{app:aux} shows that the identities of
Proposition~\ref{prop:closed} and Theorem~\ref{thm:decomp} hold at each site
separately, with $\delta_i$ evaluated in the detached backward graph.

Finally, we state the scope of this decomposition. The radial and tangential
components are an orthogonal decomposition with respect to the direction of
$\WQ$, and analyzing scale-invariant weights in terms of norm and angle is
established practice \citep{vanlaarhoven2017,wan2021,kosson2024}. The
decomposition and Theorem~\ref{thm:decomp} describe exactly where the gradient
difference goes in a single SGD step; for other optimizers they describe the
gradient that enters the optimizer. With more elaborate optimizers, the effect
of each component on the final loss requires a separate analysis.

\section{Consequences of the Radial Component Under AdamW}
\label{sec:adamw}

The results of Sections~\ref{sec:closed} and~\ref{sec:decomp} are exact only
for SGD, where the radial component merely rescales $\WQ$ and leaves the
output unchanged. Transformers, however, are commonly trained with AdamW,
which applies weight decay and rescales the update of each coordinate
separately, so that the size of the update is largely independent of the norm
of $\WQ$. The norm of $\WQ$ then affects the angle by which it rotates at each
step, and the radial component affects training indirectly. This section
gives an approximate analysis under AdamW and derives two predictions that can
be tested in training: the detached $\WQ$ approaches a larger equilibrium norm
(defined below), and its angular update per step becomes smaller.

One AdamW step \citep{kingma2015,loshchilov2019} on $\WQ$ is
\begin{equation*}
\WQ^{+} = (1 - \eta\lambda)\, \WQ - \eta\, U, \qquad
U = \hat m \oslash \bigl(\sqrt{\hat v} + \epsilon_{\mathrm A}\bigr),
\end{equation*}
where $\lambda$ is the weight-decay coefficient, $\hat m$ and $\hat v$ are the
bias-corrected exponential moving averages of the gradient and of its
elementwise square, division and square root are elementwise, and
$\epsilon_{\mathrm A}$ is the stability constant of Adam. $U$ is not the
gradient. We only use two scalars of $U$,
\begin{equation*}
u = \norm{U}, \qquad c = -\frac{\ip{U}{\WQ}}{\norm{U}\, \norm{\WQ}} .
\end{equation*}
Here $u$ is the size of the update and $c$ is the cosine of the angle between
the actual update direction $-U$ and $\WQ$, positive when the update points
outward. Expanding $\norm{\WQ^{+}}^2$ and dropping terms of order
$\eta^2\lambda$,
\begin{equation*}
\norm{\WQ^{+}}^2 - \norm{\WQ}^2 \approx
 -2\eta\lambda\, \norm{\WQ}^2 + 2\eta\, c\, u\, \norm{\WQ} + \eta^2 u^2 .
\end{equation*}
The three terms on the right are, in order, the shrinkage due to weight decay,
the growth due to the part of the update along $\WQ$, and the second-order
growth due to the size of the update itself. If $u$ and $c$ are roughly
constant over a period, setting the right-hand side to zero means that the
shrinkage from weight decay exactly offsets the growth from the update, and
$\norm{\WQ}$ is unchanged by a step. We call the norm that satisfies this
condition the equilibrium norm of $\WQ$ \citep{kosson2024}; it is
\begin{equation}
\norm{\WQ}_{\mathrm{eq}}
 = \frac{u\,\bigl(c + \sqrt{c^2 + 2\eta\lambda}\bigr)}{2\lambda} .
\label{eq:eq}
\end{equation}
The time constant for approaching it is about $1/(\eta\lambda)$ steps. The
angular update per step is approximately the part of the update orthogonal to
$\WQ$ divided by the norm, $\eta u\sqrt{1 - c^2}/\norm{\WQ}$.

The equilibrium norm takes a simple form in two limiting regimes. When
$c \ll \sqrt{2\eta\lambda}$, the equilibrium norm is about
$u\sqrt{\eta/(2\lambda)}$ and the angular update per step is about
$\sqrt{2\eta\lambda}$, independent of both $c$ and $u$; this is the rotational
equilibrium of \citet{kosson2024}. When $c \gg \sqrt{2\eta\lambda}$, the
equilibrium norm is about $cu/\lambda$ and the angular update per step is
about $\eta\lambda/c$, which decreases as $c$ grows.

The equilibrium norm is connected to the gradient difference through $c$. By
Theorem~\ref{thm:decomp}(i), the gradient under the standard rule is
orthogonal to $\WQ$, so $c$ arises only from the deviation introduced by the
elementwise rescaling of Adam. Under the detached rule, the component of the
gradient along $\WQ$ is $\sum_i \ip{\qh_i}{\delta_i}/\norm{\WQ}$. When the
head as a whole favors more concentrated attention, a step along the negative
gradient has an outward component, and $c$ increases accordingly. By
equation~\eqref{eq:eq}, the detached $\WQ$ then equilibrates at a larger norm.
If the standard runs are in the first regime and the detached runs in the
second, the ratio of their angular updates per step is about
$\sqrt{\eta\lambda/2}/c$. For the recipe used in our experiments, with peak
learning rate 0.0024 and weight decay 0.1, the threshold
$\sqrt{2\eta\lambda}$ is 0.022 and the time constant is about 4{,}167 steps.

\section{Experiments}
\label{sec:exp}

This section tests the conclusions of the preceding sections on a small
Transformer language model. The subsections address the following questions
in turn: whether detaching the denominator incurs a loss cost
(Section~\ref{sec:exp-loss}); how large the three components of the gradient
difference are during training (Section~\ref{sec:exp-comp}); whether the norm
growth and the smaller angular update predicted in Section~\ref{sec:adamw}
occur (Section~\ref{sec:exp-geom}); whether the cost depends on the choice of
learning rate and on a learnable scale (Section~\ref{sec:exp-robust}); and
whether the norm growth is accompanied by the cost
(Section~\ref{sec:exp-interv}). The main text keeps only the data that
support the conclusions; complete numbers and experimental details are given
in Appendix~\ref{app:details}.

\subsection{Setup}
\label{sec:exp-setup}

We train a decoder-only Transformer language model \citep{vaswani2017} with
about 17.65 million parameters. Queries and keys are each RMS-normalized
without a learnable scale, as in Section~\ref{sec:setup}. The training data
is drawn from FineWeb-Edu \citep{penedo2024}, and the evaluation metric is
the mean negative log-likelihood per token on held-out data, in nats.
WikiText-103 \citep{merity2017} is not used for training and serves for
evaluation only.

For each random seed, the same model is trained once under each of three
backward rules: S is the standard rule, QD detaches the query denominator in
all layers, and KD detaches the key denominator in all layers. The three
share the same forward computation, initialization and data order. We use the
AdamW optimizer \citep{loshchilov2019} with $\beta_1 = 0.9$,
$\beta_2 = 0.95$ and weight decay 0.1. Training runs for 21{,}400 steps with
warmup followed by cosine decay of the learning rate; the peak learning rate
0.0024 is selected under the standard rule and shared by the three rules.
Unless otherwise specified, each comparison uses six random seeds paired by
seed, and we report the mean of the paired differences with a 95\% confidence
interval. The S and QD runs were repeated once on different hardware (the
replication), and the diagnostics recorded during training come from this
replication. The model architecture, the remaining hyperparameters and the
statistical procedure are described in Appendix~\ref{app:details}.

\subsection{Loss Cost of Detaching the Denominator}
\label{sec:exp-loss}

Proposition~\ref{prop:closed}(c) shows that every update under the detached
rule carries an extra right factor $I - \eta M$ relative to the standard rule,
while the forward computation of the two rules is identical. Here we test
whether this difference, accumulated over training, changes the outcome. We
find that it raises the test loss by about 0.06 nats per token, which
corresponds to an increase in perplexity of about 6\%, with the same sign in
all six seeds (Table~\ref{tab:loss}). The gap is unchanged when training is
repeated on different hardware, and it is also present on WikiText-103, which
is not used for training. Section~\ref{sec:decomp} notes that key-side
detaching has the same structure as query-side detaching, and detaching the
key denominator incurs a similar cost in the experiment.

\begin{table}[t]
\centering
\caption{Paired differences in test negative log-likelihood at the end of
training (six seeds). The remaining comparisons are given in
Appendix~\ref{app:details}.}
\label{tab:loss}
\small
\begin{tabular}{llcc}
\toprule
Dataset & Comparison & Mean & 95\% CI \\
\midrule
FineWeb-Edu & QD $-$ S & 0.0600 & $[0.0546,\ 0.0653]$ \\
FineWeb-Edu & QD $-$ S (replication) & 0.0596 & $[0.0544,\ 0.0648]$ \\
FineWeb-Edu & KD $-$ S & 0.0528 & $[0.0497,\ 0.0559]$ \\
WikiText-103 & QD $-$ S & 0.0732 & $[0.0505,\ 0.0959]$ \\
\bottomrule
\end{tabular}
\end{table}

\begin{figure}[t]
\centering
\includegraphics[width=0.68\linewidth]{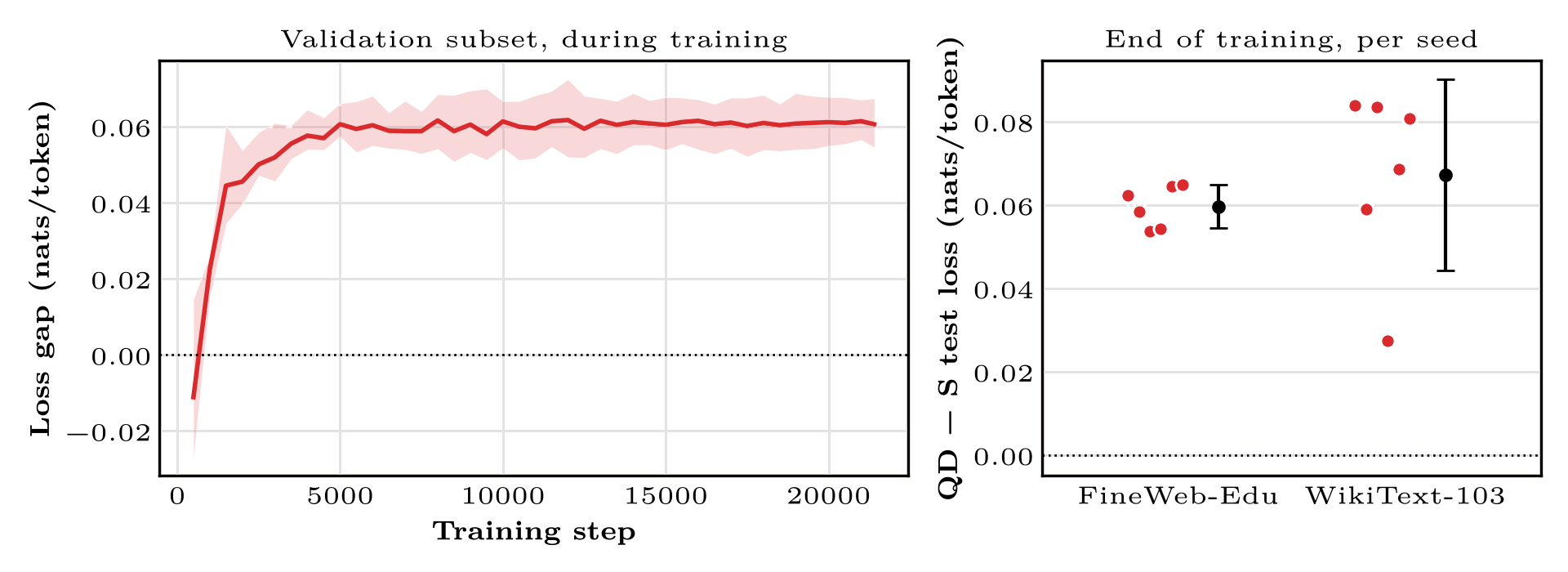}
\caption{Loss gap between QD and S. Left: gap on the validation subset over
training; the curve is the mean over six seeds and the shaded band is the
min--max range. Right: test-loss gap at the end of training; red dots are
individual seeds and black shows the mean and 95\% confidence interval
(replication, six seeds).}
\label{fig:loss}
\end{figure}

The loss gap forms early in training and then stays stable
(Figure~\ref{fig:loss}): it reaches its final value within the first 5{,}000
steps and does not change afterwards.

\subsection{Size of the Three Components of the Gradient Difference}
\label{sec:exp-comp}

Theorem~\ref{thm:decomp} splits the gradient difference into radial,
tangential and upstream components, and equation~\eqref{eq:radial} gives the
coefficient of the radial one. Here we measure these quantities directly
during training. We find that equation~\eqref{eq:radial} holds along the
training trajectory and that none of the three components is negligible at
any point in training, the tangential component being the largest.

Equation~\eqref{eq:radial} states that the inner product of the gradient with
$\WQ$ is zero under the standard rule and equals
$\sum_i \ip{\qh_i}{\delta_i}$ of the head under the detached rule. The inner
products recorded during training agree with this: the inner product is
consistently negative on the detached $\WQ$ and more than three orders of
magnitude smaller on $\WQ$ under the standard rule and on the non-detached
$\WK$ in QD, and the error in checking the third identity of
equation~\eqref{eq:radial} along the training trajectory is at the level of
float32 rounding (Appendix~\ref{app:details}). Since the inner product is
negative, Lemma~\ref{lem:beta} implies that these heads as a whole favor more
concentrated attention, so every step of the detached rule pushes $\WQ$
outward; this is the origin of the norm growth in
Section~\ref{sec:exp-geom}.

We measure the size of each component by the ratio of its norm to that of the
standard gradient (Figure~\ref{fig:comp}; the definition and the values at
selected steps are given in Appendix~\ref{app:details}). All three ratios
exceed 1\% for all seeds and checkpoints. The tangential component is the
largest, about half of the standard gradient at the end of training; the
upstream component follows at about 0.1 to 0.2; the radial component is the
smallest, below one tenth. By Theorem~\ref{thm:decomp}(ii), the tangential
component is the part of the gradient difference that changes the query
directions of the tokens, so the detached rule acts on $\WQ$ mainly through
the query directions rather than through the norm of $\WQ$.

\begin{figure}[t]
\centering
\includegraphics[width=0.75\linewidth]{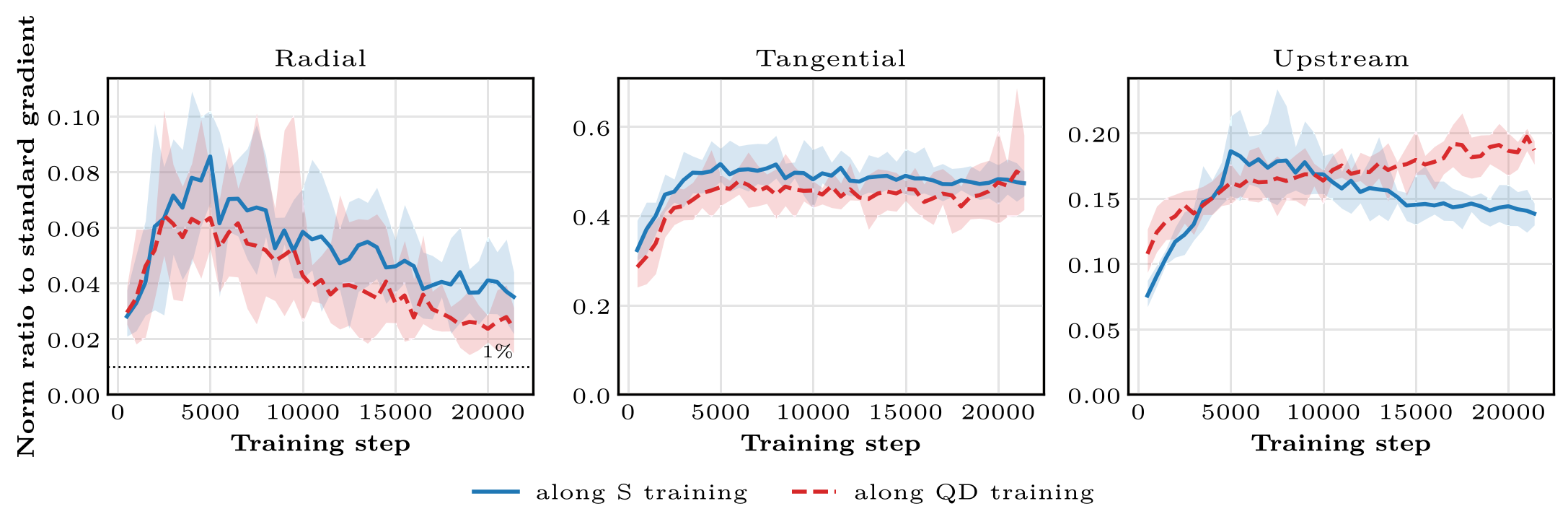}
\caption{Norm ratios of the radial, tangential and upstream components of the
gradient difference to the standard gradient over training, computed along the
S and QD trajectories. Curves are means over six seeds, shaded bands are the
min--max range, and the dotted line marks 1\%.}
\label{fig:comp}
\end{figure}

\subsection{Norm Growth and Angular Update on the Detached Side}
\label{sec:exp-geom}

\begin{figure}[t]
\centering
\includegraphics[width=0.8\linewidth]{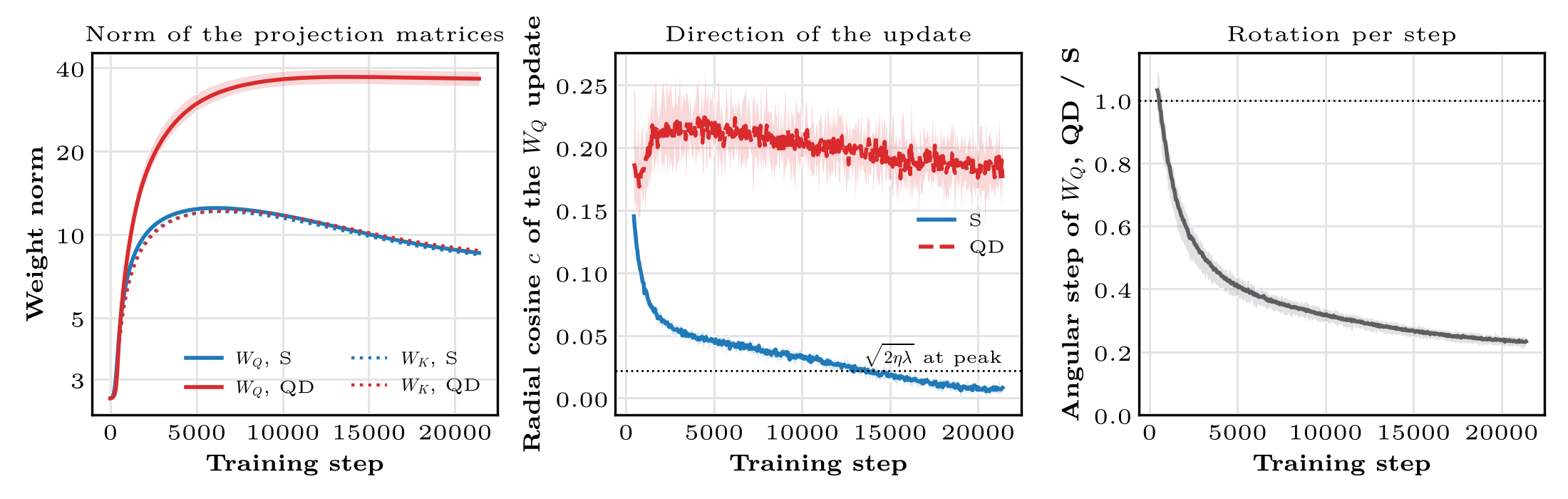}
\caption{Left: norms of $\WQ$ and $\WK$ over training (log scale). Middle:
cosine $c$ between the actual update direction of $\WQ$ and $\WQ$; the dotted
line is the threshold $\sqrt{2\eta\lambda}$ at the peak learning rate. Right:
ratio of the angular update per step of $\WQ$ in QD to that in S. Curves are
means over six seeds and shaded bands are the min--max range (replication).}
\label{fig:geom}
\end{figure}

Section~\ref{sec:adamw} derives two predictions from
equation~\eqref{eq:radial} and the equilibrium norm~\eqref{eq:eq}: the
detached projection matrix equilibrates at a larger norm, and its angular
update per step becomes smaller. We find that both occur in training, and
only on the detached side (Table~\ref{tab:geom}, Figure~\ref{fig:geom}).

\begin{table}[t]
\centering
\caption{Geometric quantities for S and QD (replication, six seeds). Norms,
query collinearity and stable rank are taken at the end of training; $c$ and
the angular update per step are averaged after warmup. Ratios are QD over S.}
\label{tab:geom}
\small
\begin{tabular}{lccl}
\toprule
Quantity & S & QD & QD versus S \\
\midrule
Norm of $\WQ$ & 8.6 & 36.5 & ratio 4.26 $[4.03,\ 4.50]$ \\
Norm of $\WK$ & -- & -- & ratio 1.036 $[1.020,\ 1.052]$ \\
Radial cosine of the update, $c$ & 0.033 & 0.200 &
 difference 0.167 $[0.157,\ 0.177]$ \\
Angular update per step & -- & -- & ratio 0.48 $[0.45,\ 0.50]$ \\
Query collinearity & 0.289 & 0.553 & difference 0.263 $[0.239,\ 0.288]$ \\
Stable rank of $\WQ$ & 9.11 & 4.73 &
 difference $-4.37$ $[-4.94,\ -3.80]$ \\
\bottomrule
\end{tabular}
\end{table}

In QD, $\WQ$ becomes markedly longer, while the non-detached $\WK$ of the same
runs is almost unchanged; with key-side detaching the roles are exchanged
(Appendix~\ref{app:details}). Equation~\eqref{eq:eq} attributes this growth
to $c$: by equation~\eqref{eq:radial}, the gradient under the detached rule
has a component along $\WQ$, which raises the cosine $c$ between the actual
update direction and $\WQ$. The measured $c$ in QD is about nine times the
threshold $\sqrt{2\eta\lambda}$, whereas in S it falls below the threshold as
training proceeds; the size of the update $u$ is almost the same in the two.
QD is therefore in the second regime of Section~\ref{sec:adamw}, with
equilibrium norm about $cu/\lambda$. Substituting the measured $u$ and $c$
into equation~\eqref{eq:eq} gives a predicted norm that is within 10\% of the
measured norm of QD from the middle of training onward
(Appendix~\ref{app:details}).

As $\WQ$ grows while $u$ stays the same, the angular update per step,
$\eta u\sqrt{1 - c^2}/\norm{\WQ}$, decreases: in QD it falls to about half of
its value in S and declines further over training (Figure~\ref{fig:geom},
right).

The norm growth and the loss gap follow different time courses. The loss gap
has reached its final value within the first 5{,}000 steps
(Section~\ref{sec:exp-loss}), whereas the norm ratio of $\WQ$ keeps
increasing after that point (Figure~\ref{fig:geom}, left). The loss gap
therefore does not track the norm ratio; Section~\ref{sec:exp-interv} tests
the relation between the two with interventions.

The query directions on the detached side also change.
Corollary~\ref{cor:rot} shows that when the coefficient is negative and the
inputs are positively correlated, the gradient difference pulls $q_j$ toward
$q_i$, so the query directions of different tokens tend to align. We quantify
this with two measures: query collinearity is the squared norm of the mean of
the unit queries of all tokens in a fixed batch, and equals 1 when all queries
point in the same direction; the stable rank of $\WQ$ is $\norm{\WQ}^2$
divided by the square of its largest singular value \citep{rudelson2007}. At
the end of training, the query collinearity of QD is about twice that of S
and the stable rank of $\WQ$ about half (Table~\ref{tab:geom}); that is, the
query directions are more concentrated and the effective dimension of $\WQ$
is lower.

\subsection{Robustness to the Learning Rate and a Learnable Scale}
\label{sec:exp-robust}

The cost reported above was measured at a learning rate selected under the
standard rule and for QK normalization without a learnable scale. Here we vary
these two conditions separately. We find that the cost does not disappear
with the choice of learning rate, but that it is not resolved for the variant
with a learnable scale.

\begin{figure}[t]
\centering
\includegraphics[width=0.62\linewidth]{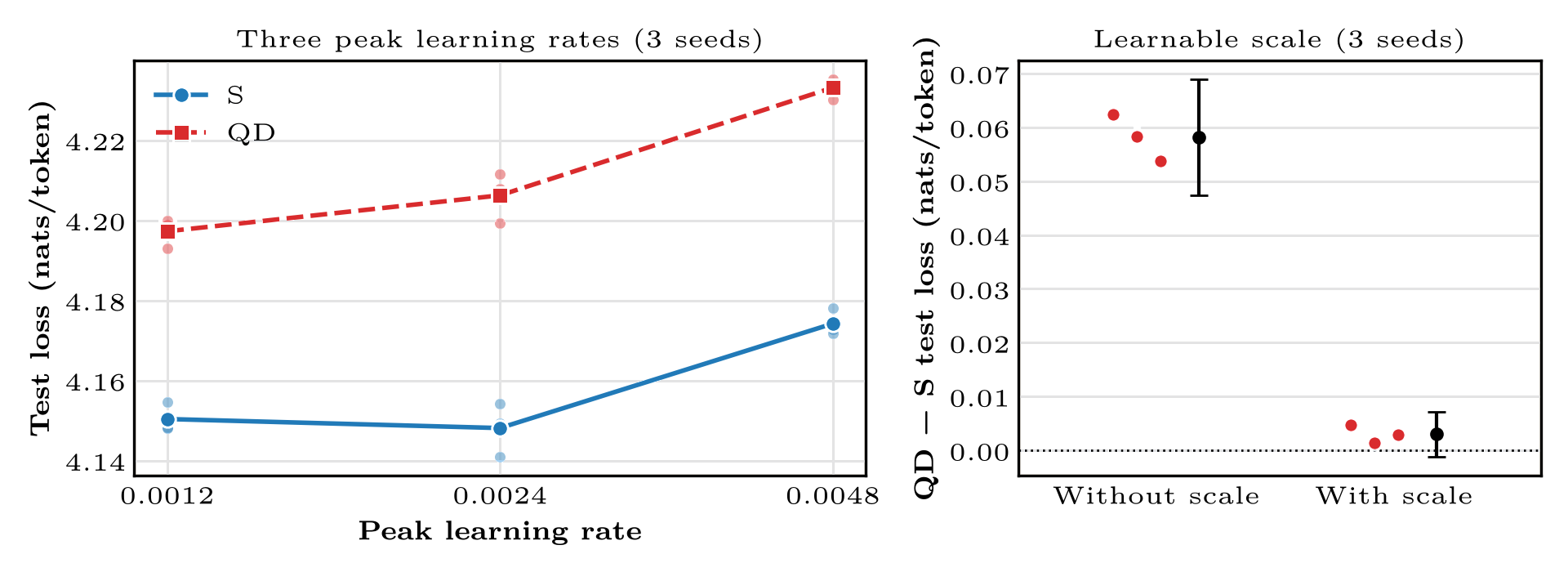}
\caption{Left: test loss of S and QD at three peak learning rates; light
points are individual seeds and dark markers are means over three seeds.
Right: test-loss difference QD $-$ S without and with a learnable scale; red
dots are individual seeds and black shows the mean and 95\% confidence
interval (three seeds each).}
\label{fig:lr}
\end{figure}

Since the peak learning rate was selected under the standard rule only, the
cost could simply reflect the fact that the learning rate was not tuned
separately for the detached rule. We train S and QD at three peak learning
rates (0.0012, 0.0024 and 0.0048) with three seeds each
(Figure~\ref{fig:lr}, left). QD is worse than S at all three learning rates.
When each rule uses the learning rate with the best validation loss, the test
losses still differ by 0.049 $[0.028, 0.070]$. The cost cannot be explained
by the choice of learning rate.

The situation differs with a learnable scale. A common form of QK
normalization multiplies by a learnable scalar $g$ after normalizing
\citep{henry2020}. In that case the radial inner product of
equation~\eqref{eq:radial} equals $g\,\partial L/\partial g$
(Appendix~\ref{app:gain}), so the radial component is tied to the gradient of
the scale. We multiply the normalized queries of each head by a learnable
scalar initialized to 1 and again compare S and QD, with three seeds each
(Figure~\ref{fig:lr}, right). The loss difference drops to 0.003
$[-0.001, 0.007]$, an interval that contains zero, and the norm growth and
the reduction in angular update of Section~\ref{sec:exp-geom} largely
disappear as well. The ratios of the tangential and upstream components
remain of the same order as without a scale, so the size of the components
does not by itself determine the cost. The phenomenon reported in this paper
is therefore specific to QK normalization without a learnable scale.

\subsection{Component Interventions}
\label{sec:exp-interv}

The decomposition of Theorem~\ref{thm:decomp} can be turned directly into
interventions that keep only some components of the gradient difference. We
use them to test whether the norm growth of Section~\ref{sec:exp-geom} is
accompanied by a loss cost. We find that keeping only the radial component
reproduces the norm growth and the smaller angular update, but not the loss
cost.

$\QR$ keeps only the radial component: $\WQ$ receives the standard gradient
plus the radial component, and all other parameters receive the standard
gradient. $\QTU$ keeps the tangential and upstream components: $\WQ$ receives
the detached-rule gradient minus the radial component, and all other
parameters receive the detached-rule gradient. In this notation, S keeps no
component and QD keeps all three. The gradients are modified before gradient
clipping and AdamW, the forward computation is the same as in S, and each arm
uses three seeds (Table~\ref{tab:interv}; see also Figure~\ref{fig:interv} in
Appendix~\ref{app:details}).

\begin{table}[t]
\centering
\caption{Component interventions (three seeds). Brackets give 95\% confidence
intervals; norm and angular-update ratios are relative to S and defined as in
Table~\ref{tab:geom}; query collinearity is taken at the end of training.}
\label{tab:interv}
\small
\setlength{\tabcolsep}{4pt}
\begin{tabular}{llcccc}
\toprule
Arm & Components kept & Test loss $-$ S & $\WQ$ norm ratio &
Angular-update ratio & Query collinearity \\
\midrule
S & none & 0 & 1 & 1 & 0.296 \\
$\QR$ & radial & 0.0030 $[-0.0045,\ 0.0104]$ & 3.52 & 0.52 & 0.283 \\
$\QTU$ & tangential, upstream & 0.0741 $[0.0645,\ 0.0836]$ & 0.70 & 1.25 &
 0.592 \\
QD & all three & 0.0581 $[0.0474,\ 0.0689]$ & 4.22 & 0.49 & 0.543 \\
\bottomrule
\end{tabular}
\end{table}

With the radial component alone ($\QR$), the norm growth of $\WQ$ and the
reduction in its angular update per step are close to those of QD, yet the
test loss cannot be distinguished from that of S. With the radial component
removed ($\QTU$), $\WQ$ no longer grows and the loss cost remains. Norm
growth and a smaller angular update are therefore not sufficient to produce
the cost. What the two arms with a cost (QD and $\QTU$) share is the
tangential and upstream components, together with higher query collinearity;
with a learnable scale the cost is not resolved and query collinearity does
not increase either (Appendix~\ref{app:details}).

We do not convert these results into a share of the cost per component. The
loss differences of the two intervention arms sum to more than the cost of
QD, so the effects of the components are not additive; in addition, $\QTU$
triggers gradient clipping on about a third of the training steps, and we
cannot tell how much of its loss difference is related to clipping
(Appendix~\ref{app:details}).

\section{Conclusions and Future Work}
\label{sec:conclusion}

\subsection{Conclusions}

We wrote the gradient difference caused by detaching the denominator in QK
normalization in the closed form $\WQ M$ and decomposed it exactly into
radial, tangential and upstream components. For a single token, this term
changes only the norm of the query; through the shared $\WQ$, it changes the
query directions of other tokens at first order and is passed on to earlier
layers. On a Transformer with 17.65 million parameters, detaching the query
denominator raises the test loss by 0.060 nats per token, and a replication on
different hardware gives the same result. The three components stay above 1\%
of the standard gradient throughout training, and the tangential component is
about half of the standard gradient. Norm growth and a smaller angular update
per step appear only on the detached side, and the equilibrium formula of
Section~\ref{sec:adamw} is within 10\% of the measured norm of the detached
runs in the middle and late stages of training. These results concern QK
normalization without a learnable scale.

Code and data manifests will be made publicly available.

\subsection{Future Work}

Our experiments use a single model scale, but the loss cost of detaching the
denominator recurs across two hardware setups, two evaluation datasets, three
peak learning rates, and both the query and the key side. Several directions
follow. First, the formulas of Sections~\ref{sec:closed}
and~\ref{sec:decomp} are exact for one step of gradient descent and describe
the gradient that enters the optimizer; bringing the elementwise rescaling,
momentum and gradient clipping of AdamW into the analysis would connect the
single-step structure to the loss gap accumulated over many steps, and the
norm balance of Section~\ref{sec:adamw} is a first step in this direction.
Second, norm growth and a smaller angular update alone are not accompanied by
a resolvable cost, whereas every arm with a cost also shows higher query
collinearity (Section~\ref{sec:exp-interv}). This suggests testing the
relation between the tangential and upstream components and query
collinearity, with interventions that avoid interference from gradient
clipping and handle the non-additivity of the components; detaching layer by
layer could further localize the cost. Third, the identity of
Appendix~\ref{app:gain}, which ties the radial inner product to the gradient
of the scale, is a starting point for explaining why no cost is resolved with
a learnable scale (Section~\ref{sec:exp-robust}). Finally, the derivation
uses only $q_i = \WQ x_i$ and $\qh_i = q_i/s_i$, so it applies equally to
other sites where a shared linear map is followed by norm normalization;
repeating our measurements at such sites, and with larger models, other
optimizers, rotary position embeddings and grouped-query attention, would
establish how far the phenomenon extends.

\clearpage
\appendix
\numberwithin{equation}{section}

\section{Proofs}
\label{app:proofs}

\subsection{Jacobian of the Standard Rule}
\label{app:jac}

We drop the index $i$. Taking the gradient of $s^2 = \norm{q}^2/d +
\varepsilon$ with respect to $q$ gives $2s\nabla s = 2q/d$, so the gradient of
$s$ with respect to $q$ is $q/(d\,s)$. Applying the quotient rule to
$\qh = q/s$,
\begin{equation}
J = \frac{1}{s}\, I - \frac{1}{s^2}\, q\, (\nabla_q s)^\top
  = \frac{1}{s}\Bigl(I - \frac{q q^\top}{d\, s^2}\Bigr).
\label{eq:jac-eps}
\end{equation}
For $\varepsilon = 0$ we have $d\,s^2 = \norm{q}^2$, which gives
equation~\eqref{eq:jac}. The detached rule keeps only the first term $I/s$.

\subsection{Proofs of Proposition~\ref{prop:closed} and Corollaries~\ref{cor:free}--\ref{cor:norm}}
\label{app:prop1}

\begin{proof}[Proof of Proposition~\ref{prop:closed}]
(a) The detached rule treats $s_i$ as a constant, and the standard rule uses
$J_i$ from Section~\ref{sec:setup}, which is symmetric, so
\begin{equation*}
\dn_{q_i} L = \frac{\delta_i}{s_i}, \qquad
\nabla_{q_i} L = J_i^\top \delta_i
 = \frac{\delta_i}{s_i}
 - \frac{\ip{q_i}{\delta_i}}{s_i\, \norm{q_i}^2}\; q_i .
\end{equation*}
Subtracting,
\begin{equation*}
\dn_{q_i} L - \nabla_{q_i} L
 = \frac{\ip{q_i}{\delta_i}}{s_i\, \norm{q_i}^2}\; q_i
 = \frac{\ip{\qh_i}{\delta_i}}{\norm{q_i}^2}\; q_i ,
\end{equation*}
and taking the inner product of each gradient with $q_i$,
\begin{equation*}
\ip{q_i}{\nabla_{q_i} L}
 = \frac{\ip{q_i}{\delta_i}}{s_i} - \frac{\ip{q_i}{\delta_i}}{s_i} = 0,
\qquad
\ip{q_i}{\dn_{q_i} L} = \frac{\ip{q_i}{\delta_i}}{s_i}
 = \ip{\qh_i}{\delta_i} .
\end{equation*}
(b) By $q_i = \WQ x_i$ and the chain rule,
\begin{equation*}
\nabla_{\WQ} L = \sum_i \bigl(\nabla_{q_i} L\bigr)\, x_i^\top ,
\end{equation*}
and the same holds for $\dn$. Subtracting, substituting the gradient
difference from (a), and using $q_i = \WQ x_i$ to pull $\WQ$ out of the sum,
\begin{equation*}
\dn_{\WQ} L - \nabla_{\WQ} L
 = \sum_i \frac{\ip{\qh_i}{\delta_i}}{\norm{q_i}^2}\; q_i\, x_i^\top
 = \WQ \sum_i \frac{\ip{\qh_i}{\delta_i}}{\norm{q_i}^2}\; x_i x_i^\top
 = \WQ M .
\end{equation*}
(c) Writing the gradient under the detached rule as the standard gradient plus
$\WQ M$ and substituting into one update step gives the stated update.
\end{proof}

\begin{proof}[Proof of Corollary~\ref{cor:free}]
By Proposition~\ref{prop:closed}(a),
\begin{equation*}
q_i - \eta\,\dn_{q_i} L
 = \Bigl(1 - \eta\,\frac{\ip{\qh_i}{\delta_i}}{\norm{q_i}^2}\Bigr)
   \bigl(q_i - \eta\,\nabla_{q_i} L\bigr)
 - \eta^2\,\frac{\ip{\qh_i}{\delta_i}}{\norm{q_i}^2}\,\nabla_{q_i} L .
\end{equation*}
For $\eta$ small enough, the coefficient of the first term on the right is
positive. Since $\qh_i$ is invariant to positive rescaling of $q_i$ and smooth
near $q_i \neq 0$, the $\qh_i$ after one step under the two rules differ by
$O(\eta^2)$.
\end{proof}

\begin{proof}[Proof of Corollary~\ref{cor:rot}]
By the expression for $M$ in Proposition~\ref{prop:closed}(b) and
$q_i = \WQ x_i$,
\begin{equation*}
\WQ M\, x_j
 = \sum_i \frac{\ip{\qh_i}{\delta_i}}{\norm{q_i}^2}\, \WQ x_i\, x_i^\top x_j
 = \sum_i \frac{\ip{\qh_i}{\delta_i}}{\norm{q_i}^2}\, \ip{x_i}{x_j}\, q_i ,
\end{equation*}
and multiplying by $-\eta$ gives the first identity. The expression for $J_j$
gives $J_j q_j = 0$. Applying $J_j$ to the first identity removes the term
$i = j$ and gives the second.
\end{proof}

\begin{proof}[Proof of Corollary~\ref{cor:norm}]
By Proposition~\ref{prop:closed}(c), the $q_i$ after one step under the two
rules differ by $-\eta\,\WQ M x_i$, so the squared norms differ by
\begin{equation*}
-2\eta\,\ip{q_i}{\WQ M x_i} + O(\eta^2)
 = -2\eta \sum_j \frac{\ip{\qh_j}{\delta_j}}{\norm{q_j}^2}\,
   \ip{x_i}{x_j}\, \ip{q_i}{q_j} + O(\eta^2) ,
\end{equation*}
where the second equality uses the first identity of Corollary~\ref{cor:rot}.
\end{proof}

\subsection{Proof of Theorem~\ref{thm:decomp}}
\label{app:thm1}

\begin{proof}[Proof of Theorem~\ref{thm:decomp}]
(i) From $q_i = \WQ x_i$ we have
$\ip{\WQ}{\WQ x_i x_i^\top} = \norm{q_i}^2$; substituting into the expression
for $M$ gives the first identity. By the chain rule,
\begin{equation*}
\ip{\WQ}{\nabla_{\WQ} L}
 = \sum_i \ip{\WQ}{\nabla_{q_i} L\, x_i^\top}
 = \sum_i \ip{q_i}{\nabla_{q_i} L} = 0 ,
\end{equation*}
where the last step uses the orthogonality of the standard gradient and $q_i$
from the proof of Proposition~\ref{prop:closed}. Adding the two identities and
using Proposition~\ref{prop:closed}(b) gives the third.
(ii) By the definition of the radial component,
\begin{equation*}
J_j\, (\WQ M)_{\parallel}\, x_j
 = \frac{\ip{\WQ}{\WQ M}}{\norm{\WQ}^2}\, J_j q_j = 0 .
\end{equation*}
(iii) The two rules differ only at $q_i$, and by $q_i = \WQ x_i$,
\begin{equation*}
\dn_{x_i} L - \nabla_{x_i} L
 = \WQ^\top \bigl(\dn_{q_i} L - \nabla_{q_i} L\bigr) ;
\end{equation*}
substituting Proposition~\ref{prop:closed}(a) gives the result.
\end{proof}

\subsection{Key-Side Coefficients}
\label{app:key}

Since $z_{ij} = \ip{\qh_i}{\kh_j}/\sqrt d$, the gradient of the loss with
respect to $\kh_j$ is the sum over $i$ of $\partial L/\partial z_{ij}$ times
$\qh_i/\sqrt d$, so
\begin{equation*}
\Bigl\langle \kh_j,\, \frac{\partial L}{\partial \kh_j}\Bigr\rangle
 = \sum_i z_{ij}\, \frac{\partial L}{\partial z_{ij}}, \qquad
\sum_j \Bigl\langle \kh_j,\, \frac{\partial L}{\partial \kh_j}\Bigr\rangle
 = \sum_{i,j} z_{ij}\, \frac{\partial L}{\partial z_{ij}}
 = \sum_i \ip{\qh_i}{\delta_i} .
\end{equation*}
The sums run over visible pairs $(i, j)$ only. Replacing $\WQ$, $q_i$, $x_i$,
$\delta_i$ by $\WK$, $k_j$, $x_j$, $\partial L/\partial \kh_j$ in the proofs
of Proposition~\ref{prop:closed} and Theorem~\ref{thm:decomp} gives the
corresponding key-side results.

\subsection{Proof of Proposition~\ref{prop:aux}}
\label{app:aux}

Let $F$ denote the function in brackets in Proposition~\ref{prop:aux}, with
the coefficients $\ip{\qh_j}{\delta_j}$ treated as constants and $\delta_j$
the backpropagated gradient at site $j$ in the detached backward graph.
Backpropagating $F$ under the standard rule amounts to backpropagating $L$
under the standard rule and adding, at $q_j$ for each site $j$, an extra
gradient equal to $\ip{\qh_j}{\delta_j}$ times the gradient of $\log s_j$ with
respect to $q_j$, because $\log s_j$ depends on the parameters only through
$q_j$.

Order all detached sites topologically from output to input and argue by
induction over sites. Suppose that at every site downstream of $j$ the
standard backward pass of $F$ and the detached backward pass of $L$ emit the
same signal. Both use the standard rule everywhere else, so the backpropagated
gradient reaching $\qh_j$ is $\delta_j$ in both. At site $j$, the backward
pass of $F$ passes to $q_j$
\begin{equation*}
J_j^\top \delta_j + \ip{\qh_j}{\delta_j}\, \nabla_{q_j} \log s_j
 = \frac{\delta_j}{s_j} - \frac{\ip{q_j}{\delta_j}}{d\, s_j^3}\, q_j
 + \frac{\ip{q_j}{\delta_j}}{s_j}\cdot \frac{q_j}{d\, s_j^2}
 = \frac{\delta_j}{s_j} ,
\end{equation*}
where we used equation~\eqref{eq:jac-eps} and the gradient
$q_j/(d\,s_j^2)$ of $\log s_j$ with respect to $q_j$. The right-hand side is
the signal that the detached rule passes to $q_j$. This completes the
induction, and the two gradients agree for every parameter. The argument holds
for any $\varepsilon \ge 0$. The same induction shows that, when all layers
are detached simultaneously, Proposition~\ref{prop:closed} and
Theorem~\ref{thm:decomp} hold at each site separately, provided $\delta_i$ is
evaluated in the detached backward graph. \qed

\subsection{Modifications for \texorpdfstring{$\varepsilon > 0$}{epsilon > 0}}
\label{app:eps}

By equation~\eqref{eq:jac-eps},
$1/(d\,s^2) = (1 - \varepsilon/s^2)/\norm{q}^2$. Substituting into the proof
of Proposition~\ref{prop:closed}(a) gives
\begin{equation*}
\dn_{q_i} L - \nabla_{q_i} L
 = \frac{\ip{\qh_i}{\delta_i}\,(1 - \varepsilon/s_i^2)}{\norm{q_i}^2}\; q_i,
\qquad
\ip{q_i}{\nabla_{q_i} L} = \frac{\varepsilon}{s_i^2}\, \ip{\qh_i}{\delta_i},
\qquad
\ip{q_i}{\dn_{q_i} L} = \ip{\qh_i}{\delta_i} .
\end{equation*}
Each coefficient of $M$ therefore acquires a factor $1 - \varepsilon/s_i^2$,
and Theorem~\ref{thm:decomp}(i) changes accordingly: $\ip{\WQ}{\WQ M}$ is the
sum over $i$ of $\ip{\qh_i}{\delta_i}(1 - \varepsilon/s_i^2)$, and the inner
product of the standard gradient with $\WQ$ is the sum over $i$ of
$(\varepsilon/s_i^2)\ip{\qh_i}{\delta_i}$. The inner product of the detached
gradient with $\WQ$ still equals $\sum_i \ip{\qh_i}{\delta_i}$ exactly. This
last identity holds for any $\varepsilon$, so the check in
Section~\ref{sec:exp-comp} needs no correction.

For $\varepsilon > 0$ the normalization is no longer strictly invariant to
positive rescaling, since $J q = (\varepsilon/s^2)\,\qh$. A radial step
therefore has a first-order effect on the output of order
$\varepsilon/s_i^2$, and the difference between the two $\qh_i$ in
Corollary~\ref{cor:free} acquires a first-order term of the same order. With
the $\varepsilon = 10^{-5}$ of our experiments, this term was measured at six
checkpoints and accounts for only $5.9 \times 10^{-6}$ to
$1.1 \times 10^{-5}$ of the full first-order response.

\subsection{Proof of Proposition~\ref{prop:vanish}}
\label{app:vanish}

(a) The tangential component vanishes when $\WQ M = \alpha \WQ$ with
$\alpha = \ip{\WQ}{\WQ M}/\norm{\WQ}^2$. This is equivalent to
$\WQ(M - \alpha I) = 0$. Since $M$ is symmetric, transposing gives
$(M - \alpha I)\WQ^\top = 0$, that is, $Mv = \alpha v$ for every $v$ in the
range of $\WQ^\top$, which is the row space of $\WQ$. Conversely, if $M$
equals some $\alpha'$ times the identity on the row space, then
$\WQ M = \alpha' \WQ$, which is parallel to $\WQ$, so the tangential part is
zero. Taking the inner product with $\WQ$ gives $\alpha' = \alpha$. By
equation~\eqref{eq:radial},
$\alpha = \sum_i \ip{\qh_i}{\delta_i}/\norm{\WQ}^2$.

(b) Let $h_i$ be the residual-stream vector at token $i$. An RMSNorm without
gain gives $x_i = h_i/r_i$ with $r_i = \sqrt{\norm{h_i}^2/m}$. The same
calculation as for equation~\eqref{eq:jac-eps} shows that backpropagation
through this normalization first removes the component along $x_i$ from the
gradient at $x_i$, then divides by $r_i$, and passes the result to $h_i$. By
equation~\eqref{eq:upstream}, the gradient difference at $x_i$ is a multiple
of $\WQ^\top \WQ\, x_i$, with factor $\ip{\qh_i}{\delta_i}/\norm{q_i}^2$.
When the factor is nonzero, the result is zero after removing the component
along $x_i$ if and only if $\WQ^\top \WQ\, x_i$ is parallel to $x_i$, that is,
$x_i$ is an eigenvector of $\WQ^\top \WQ$. The $h_i$ of different tokens are
distinct variables, so vanishing for all tokens is equivalent to vanishing for
each token separately. \qed

With an elementwise gain $\gamma$, we have $x_i = \Gamma h_i/r_i$, where
$\Gamma$ is the diagonal matrix with diagonal $\gamma$. The same calculation
shows that the condition becomes: $h_i/r_i$ is an eigenvector of
$\Gamma \WQ^\top \WQ\, \Gamma$. For $\gamma$ identically 1 this reduces
to (b).

\subsection{Radial Inner Product with a Learnable Gain}
\label{app:gain}

Suppose each head multiplies the normalized queries by a learnable scalar $g$,
so that $g\,\qh_i$ enters the scores, and define $\delta_i$ as the gradient of
the loss with respect to $g\,\qh_i$. The gradient with respect to $q_i$ under
the detached rule is $g\,\delta_i/s_i$, so
\begin{equation*}
\ip{\WQ}{\dn_{\WQ} L}
 = \sum_i \Bigl\langle q_i,\, \frac{g\, \delta_i}{s_i}\Bigr\rangle
 = g \sum_i \ip{\qh_i}{\delta_i}
 = g\, \frac{\partial L}{\partial g} .
\end{equation*}
This holds for any $\varepsilon$. Under gradient flow without weight decay
and with the same learning rate for $\WQ$ and $g$, the derivative of
$\norm{\WQ}^2 - g^2$ is
$-2\ip{\WQ}{\dn_{\WQ} L} + 2g\,\partial L/\partial g = 0$, so this quantity is
conserved \citep[cf.][]{kunin2021}. Discrete SGD changes it at each step by
$\eta^2$ times the difference of the squared norms of the two gradients. No
such conservation law holds under AdamW.

\section{Numerical Checks}
\label{app:numerics}

Proposition~\ref{prop:closed}, Theorem~\ref{thm:decomp},
Proposition~\ref{prop:aux} and Proposition~\ref{prop:vanish} were each checked
term by term with automatic differentiation on a small network, using the
script \texttt{qkn\_routes\_check.py}. The network contains an upstream
parameter table, a pre-RMSNorm, shared $\WQ$ and $\WK$, and causal attention.
Computations are in float64 with $\varepsilon \in \{0, 10^{-5}, 0.3\}$, for 48
checks in total. The largest relative error is $3.1 \times 10^{-16}$ for the
right-multiplication form of Proposition~\ref{prop:closed}(b),
$2.9 \times 10^{-16}$ for the radial coefficient and orthogonality in
Theorem~\ref{thm:decomp}(i), $5.0 \times 10^{-16}$ for the upstream component
of Theorem~\ref{thm:decomp}(iii) and its inner product with $x_i$,
$4.5 \times 10^{-16}$ for the key-side counterparts, and
$1.1 \times 10^{-16}$ for the isotropy conditions of
Proposition~\ref{prop:vanish}. Corollary~\ref{cor:norm} is a first-order
statement and was checked by finite differences with step $10^{-6}$; the
relative error is $6.6 \times 10^{-8}$, consistent with the remainder of the
first-order expansion. Proposition~\ref{prop:aux} was checked on a three-layer
network with all layers detached simultaneously; the two sides differ by
$4 \times 10^{-16}$.

In code, the detached rule is implemented by writing the normalization as
\texttt{q / s.detach()}; everything else is the same as under the standard
rule.

The random state used for these checks also provides one example of the
relative size of the three components, which is not representative of
training. The radial component accounts for only 9\% of the squared norm of
the gradient difference on $\WQ$ and the tangential component for 91\%; the
norm of the tangential component is 51\% of that of the standard gradient.
When $\WQ$ is chosen with orthogonal rows of equal norm, the upstream
component drops to $10^{-16}$, while the gradient difference on $\WQ$ in the
same state is still 17\% of the standard gradient.

\section{Experimental Details}
\label{app:details}
\renewcommand{\thetable}{C.\arabic{table}}
\renewcommand{\theHtable}{C.\arabic{table}}
\setcounter{table}{0}
\renewcommand{\thefigure}{C.\arabic{figure}}
\renewcommand{\theHfigure}{C.\arabic{figure}}
\setcounter{figure}{0}

Throughout this appendix, the \emph{original runs} are the first set of
training runs and the \emph{replication runs} are the second set, referred to
as the replication in the main text.

\paragraph{Model and hyperparameters.}
The model has 6 layers, width 256, 4 heads per layer with head dimension 64, a
feed-forward width of 4 times the model width, context length 256 and
vocabulary size 50{,}257, for 17{,}653{,}248 parameters in total. RMSNorm is
applied before each sublayer, position embeddings are learned, linear layers
have no bias, and the input and output embeddings are tied. The RMS
normalization of queries and keys uses $\varepsilon = 10^{-5}$. The training
data is the sample-10BT subset of FineWeb-Edu, of which 351 million tokens are
used for training and 5 million tokens each are held out for validation and
test. AdamW uses $(\beta_1, \beta_2) = (0.9, 0.95)$, stability constant
$10^{-8}$, weight decay 0.1 and global gradient clipping at 1.0. Each step
uses $64 \times 256$ tokens, with 428 warmup steps followed by cosine decay to
10\% of the peak. The two sets comprise 54 main training runs (18 original
and 36 replication), all of which completed; the original set was preceded by
10 single-seed learning-rate pilot runs.

\paragraph{Statistics.}
Intervals are computed from the $t$ distribution. The primary endpoint was
fixed before training: the difference in loss between QD and S on the
FineWeb-Edu test set at the end of training. All other results are
descriptive, and their intervals are not corrected for multiple comparisons.
The replication runs repeat S and QD of the original runs.

\paragraph{Hardware and run time.}
The original runs used one RTX 4090~D; the 18 main runs and the 10 pilot runs
took about 14 hours in total. The replication runs used
one A800 (80\,GB) with PyTorch 2.5.1 and CUDA 12.4; the 36 runs were executed
two at a time and took about 20 hours in total, a single run taking about
0.75 hours when run alone. Both sets use BF16 mixed precision, with each step
of $64 \times 256$ tokens accumulated over two micro-batches.

\paragraph{Frozen record.}
The SHA-256 hashes of the protocol, configuration and code of the replication
runs were written to a frozen record before the first run (protocol
\texttt{qkn-stage2-20261112-v1}, 12 November 2026, 13:40 UTC), and the launch
script verifies the three hashes at every start. The protocol specifies that
each (rule, seed, learning rate) combination is trained once, that the
behavior of a run is not grounds for rerunning it, and that no arms, seeds or
learning rates are added after results are seen. All 36 runs completed in a
single attempt, without restarts or numerical failures.

\paragraph{Implementation checks.}
Before training, we checked the agreement of each backward rule with the
closed forms on a model of the same size with randomly perturbed
initialization (GPU, float32). The largest relative error over all identities
is $1.2 \times 10^{-6}$. For the identity stating that, with all layers
detached, the inner product of the detached gradient with $\WQ$ equals
$\sum_i \ip{\qh_i}{\delta_i}$, the error is $3.1 \times 10^{-7}$, with
$\delta_i$ evaluated in the detached backward graph. The forward losses of the
rules under BF16 are bitwise identical.

\paragraph{Checks along the training trajectory.}
For the check in Section~\ref{sec:exp-comp}, every 500 steps we compute the
gradients on a fixed batch under each rule (float32) and compare the inner
product of the detached gradient with $\WQ$ against
$\sum_i \ip{\qh_i}{\delta_i}$ computed under the standard rule. The two should
agree on the last layer, because no detached normalization follows it and
$\delta_i$ is the same under both rules. Over the 43 checkpoints of all 36
replication runs, the relative error summed over the four heads of the last
layer is at most $8.0 \times 10^{-7}$, with median $1.0 \times 10^{-7}$. For
the other layers, $\delta_i$ must be evaluated in the detached backward graph
(Appendix~\ref{app:aux}), and this quantity was not recorded during training.

\paragraph{Diagnostics.}
The component norm ratios of Section~\ref{sec:exp-comp} and the geometric
quantities of Section~\ref{sec:exp-geom} are computed every 500 steps on 16
fixed windows of the validation set (256 tokens each), in float32 without
mixed precision, and without modifying the parameters, the optimizer state or
the random-number state. The quantities $u$ and $c$ and the per-head norms of
Section~\ref{sec:exp-geom} are computed every 50 steps from the actual
training updates.

\paragraph{Additional numbers.}
\emph{Section~\ref{sec:exp-loss}.} On the validation subset (1 million tokens,
evaluated every 500 steps), QD $-$ S is $-0.011$ $[-0.026, 0.004]$ at step
500, 0.022 at step 1{,}000, 0.046 at step 2{,}000 and 0.061 at step 5{,}000,
and stays between 0.058 and 0.062 afterwards. In the original runs, the gap of
KD reaches 0.065 at step 1{,}750 and falls back to 0.053 at the end.

\emph{Section~\ref{sec:exp-comp}.} The mean inner product of gradient and
weight after warmup is $-2.42 \times 10^{-3}$ for $\WQ$ in QD and
$-7.9 \times 10^{-7}$ for $\WQ$ in S in the replication runs; in the original
runs the two values are $-2.41 \times 10^{-3}$ and $-8.5 \times 10^{-7}$, and
it is $-2.29 \times 10^{-3}$ for $\WK$ in KD. For the non-detached $\WK$ in QD
it is $-1.2 \times 10^{-7}$. Over both trajectories, all seeds and all
checkpoints, the ratio is between 0.014 and 0.11 for the radial component,
between 0.24 and 0.69 for the tangential component, and between 0.07 and 0.23
for the upstream component. In squared norm, the radial component is on
average at most 7\% of the gradient difference on $\WQ$ across layers at steps
5{,}000 and 21{,}400. The tangential ratio is larger in the first three
layers, about 0.55 to 0.60 at the end of QD training, and about 0.28 in the
last two layers.

\emph{Section~\ref{sec:exp-geom}.} In the original runs, the norm ratio in QD
is 4.34 for $\WQ$ and 1.035 for $\WK$. The per-layer norm ratios of $\WQ$ in
QD are 1.85, 2.06, 7.33, 7.80, 8.02, 3.41 in the replication runs and 1.75,
2.27, 7.69, 8.73, 7.63, 3.27 in the original runs. The norm ratio of $\WQ$ is
2.40 at step 5{,}000, 3.10 at step 10{,}000 and 4.26 at the end of training;
after step 10{,}000 the norm of $\WQ$ (geometric mean) goes from 36.4 to 36.5
in QD and from 11.7 down to 8.6 in S. The cosine $c$ of $\WQ$ falls from 0.09
to 0.01 over training in S and stays between 0.19 and 0.21 throughout in QD.
The ratio of angular updates per step is 0.83 over the first 2{,}000 steps and
0.25 over the last 5{,}400 steps; in the three middle layers it is about 0.13
to 0.16 in the second half of training. The paired difference in key
collinearity is 0.100 $[0.085, 0.114]$.

\emph{Section~\ref{sec:exp-robust}.} In the pilot (single seed, validation
set), the validation loss of QD at learning rates 0.0012, 0.0006 and 0.0003 is
4.246, 4.279 and 4.413; QD is worse than S at all five learning rates, and the
gap between the best of each is 0.049. The cost with a scale is lower than the
cost without a scale on the same three seeds by 0.0551 $[0.0458, 0.0645]$. On
WikiText-103, QD $-$ S with a scale is $-0.0239$ $[-0.0443, -0.0036]$. With a
scale, the norm ratio of $\WK$ is 1.00 $[0.98, 1.02]$, and $c$ of $\WQ$ is
0.039 in QD and 0.034 in S; the radial, tangential and upstream ratios at the
end of training are 0.013, 0.23 and 0.14; the learned $g$ averages about 2.8
in both arms; and the difference in query collinearity, QD minus S, is
$-0.043$ $[-0.053, -0.034]$. The fraction of steps after warmup on which
gradient clipping is triggered is 1.1\% for KD, 0.09\% for QD and 0 for S in
the original runs; in the replication runs it is 0.02\% for QD, and S triggers
clipping on a single step across the six runs. At peak learning rate 0.0012 it
is 1.6\% for QD.

\emph{Section~\ref{sec:exp-interv}.} The loss difference between QD and
$\QTU$ is $-0.0159$ $[-0.0337, 0.0018]$. The clipping fractions of the three
seeds of $\QTU$ are 39\%, 19\% and 38\%.

\paragraph{Numbers not listed in the main text.}
\emph{Section~\ref{sec:exp-geom}.} With key-side detaching (KD), the norm
ratio is 3.43 for $\WK$ and 1.086 for $\WQ$.
\emph{Section~\ref{sec:exp-robust}.} With a learnable scale, the norm ratio
of $\WQ$ is 1.10 $[1.04, 1.15]$ and the ratio of angular updates per step is
0.91 $[0.89, 0.94]$.
\emph{Section~\ref{sec:exp-interv}.} The loss differences of the two
intervention arms sum to 0.0189 $[0.0036, 0.0342]$ more than the cost of QD.
The fraction of steps after warmup on which gradient clipping is triggered is
0.002\% for S and $\QR$, 0.03\% for QD and 32\% for $\QTU$. The tables moved
out of the main text are given below
(Tables~\ref{tab:loss-full}--\ref{tab:lr}).

\begin{table}[h]
\centering
\caption{All paired differences in test negative log-likelihood at the end of
training (step 21{,}400), six seeds.}
\label{tab:loss-full}
\small
\begin{tabular}{llcc}
\toprule
Dataset & Comparison & Mean & 95\% CI \\
\midrule
FineWeb-Edu & QD $-$ S & 0.0600 & $[0.0546,\ 0.0653]$ \\
FineWeb-Edu & KD $-$ S & 0.0528 & $[0.0497,\ 0.0559]$ \\
FineWeb-Edu & QD $-$ KD & 0.0071 & $[0.0015,\ 0.0128]$ \\
FineWeb-Edu & QD $-$ S (replication) & 0.0596 & $[0.0544,\ 0.0648]$ \\
\midrule
WikiText-103 & QD $-$ S & 0.0732 & $[0.0505,\ 0.0959]$ \\
WikiText-103 & KD $-$ S & 0.0721 & $[0.0431,\ 0.1011]$ \\
WikiText-103 & QD $-$ KD & 0.0011 & $[-0.0242,\ 0.0264]$ \\
WikiText-103 & QD $-$ S (replication) & 0.0672 & $[0.0442,\ 0.0901]$ \\
\bottomrule
\end{tabular}
\end{table}

\begin{table}[h]
\centering
\caption{Norm ratios of the three components of the gradient difference to the
standard gradient. The radial and tangential components are compared with the
standard gradient on $\WQ$ of each layer; the upstream component is the
gradient difference on all parameters other than $\WQ$ and is compared with
the standard gradient on those parameters. Values at each step are means over
six seeds; the minimum is taken over all seeds and checkpoints.}
\label{tab:comp}
\small
\begin{tabular}{llccccc}
\toprule
Component & Trajectory & Step 500 & Step 5{,}000 & Step 10{,}000 &
Step 21{,}400 & Minimum \\
\midrule
Radial & S & 0.028 & 0.086 & 0.058 & 0.035 & 0.021 \\
Radial & QD & 0.029 & 0.063 & 0.043 & 0.024 & 0.014 \\
Tangential & S & 0.323 & 0.516 & 0.482 & 0.473 & 0.301 \\
Tangential & QD & 0.285 & 0.464 & 0.457 & 0.488 & 0.240 \\
Upstream & S & 0.076 & 0.186 & 0.168 & 0.138 & 0.067 \\
Upstream & QD & 0.108 & 0.162 & 0.164 & 0.187 & 0.093 \\
\bottomrule
\end{tabular}
\end{table}

\begin{table}[h]
\centering
\caption{Ratio of the norm of $\WQ$ predicted by equation~\eqref{eq:eq} to the
measured norm, median over heads (replication, six seeds). The prediction
substitutes the averages of $u$, $c$ and the learning rate over the preceding
1{,}000 steps into equation~\eqref{eq:eq}.}
\label{tab:pred}
\small
\begin{tabular}{lcccc}
\toprule
Trajectory & Step 5{,}000 & Step 10{,}000 & Step 15{,}000 & Step 21{,}400 \\
\midrule
QD & 1.42 & 1.09 & 1.01 & 0.93 \\
S & 1.10 & 0.86 & 0.60 & 0.28 \\
\bottomrule
\end{tabular}
\end{table}

For QD, the prediction is within 10\% of the measurement from the middle of
training onward. For S, the ratio keeps falling. Learning-rate decay lowers
the equilibrium norm of S, while the time constant $1/(\eta\lambda)$ for
approaching equilibrium grows from 4{,}167 to 41{,}667 steps, so the measured
norm cannot keep up. The equilibrium norm of QD is approximately
$uc/\lambda$, which does not depend on the learning rate.

\begin{table}[h]
\centering
\caption{S and QD at three peak learning rates (three seeds). Validation
losses are seed means and are used to select the learning rate for each
rule.}
\label{tab:lr}
\small
\begin{tabular}{ccccc}
\toprule
Peak learning rate & Val.\ loss, S & Val.\ loss, QD &
QD $-$ S test loss & 95\% CI \\
\midrule
0.0012 & 4.2035 & 4.2500 & 0.0469 & $[0.0364,\ 0.0574]$ \\
0.0024 & 4.2011 & 4.2592 & 0.0581 & $[0.0474,\ 0.0689]$ \\
0.0048 & 4.2281 & 4.2863 & 0.0589 & $[0.0534,\ 0.0645]$ \\
\bottomrule
\end{tabular}
\end{table}

\begin{figure}[h]
\centering
\includegraphics[width=0.5\linewidth]{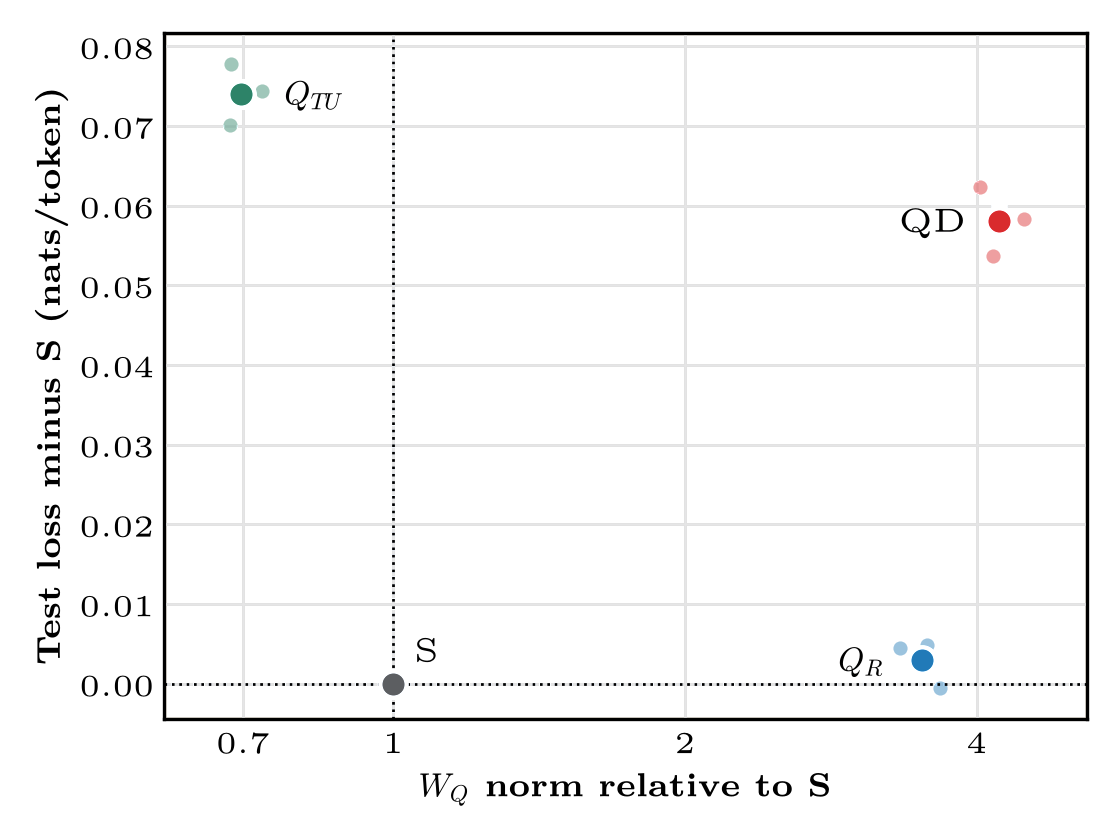}
\caption{Test-loss difference relative to S against the norm ratio of $\WQ$
for each arm of Section~\ref{sec:exp-interv} (log scale on the horizontal
axis). Light points are individual seeds and dark points are means over three
seeds.}
\label{fig:interv}
\end{figure}

\end{document}